\documentclass{article}

\usepackage{microtype}
\usepackage{graphicx}
\usepackage{subfigure}
\usepackage{booktabs} 

\usepackage{comment}
\usepackage{amsmath,amssymb}
\usepackage{color}
\usepackage{adjustbox}
\usepackage{natbib}

\usepackage{hyperref}

\usepackage{algorithm}
\usepackage{algorithmic}

\floatstyle{ruled}
\newfloat{algorithm}{tbp}{loa}
\renewcommand{\algorithmiccomment}[1]{/* #1 */}
\renewcommand{\algorithmicforall}{\textbf{for each}}

\newcommand{\din}{\mathcal D}

\usepackage[accepted]{icml2021}
\icmltitlerunning{Immuno-mimetic Deep Neural Networks (Immuno-Net)}

\begin{document}

\twocolumn[
\icmltitle{Immuno-mimetic Deep Neural Networks (Immuno-Net)  
}



\icmlsetsymbol{equal}{*}

\begin{icmlauthorlist}
\icmlauthor{Ren Wang}{to}
\icmlauthor{Tianqi Chen}{to}
\icmlauthor{Stephen Lindsly}{to}
\icmlauthor{Cooper Stansbury}{to}
\icmlauthor{Indika Rajapakse}{to}
\icmlauthor{Alfred Hero}{to}
\end{icmlauthorlist}

\icmlaffiliation{to}{University of Michigan, Michigan, United States}

\icmlcorrespondingauthor{Alfred Hero}{hero@eecs.umich.edu}
\icmlcorrespondingauthor{Indika Rajapakse}{indikar@umich.edu}
\icmlcorrespondingauthor{Ren Wang}{renwang@umich.edu}

\icmlkeywords{Machine Learning, ICML }

\vskip 0.3in
]



\printAffiliationsAndNotice{}  


\begin{abstract}
Biomimetics has played a key role in the evolution of artificial neural networks. Thus far, {\em in silico} metaphors have been dominated by concepts from neuroscience and cognitive psychology. In this paper we introduce a different type of biomimetic model, one that borrows concepts from the immune system, for designing robust deep neural networks. This immuno-mimetic model leads to a new computational biology framework for robustification of deep neural networks against adversarial attacks. Within this Immuno-Net framework we define a robust adaptive immune-inspired learning system (Immuno-Net RAILS) that emulates, {\em in silico}, the  
adaptive biological mechanisms of B-cells that are used to defend a mammalian host against pathogenic attacks. 
When applied to image classification tasks on  benchmark datasets, we demonstrate that Immuno-net RAILS results in improvement of as much as $12.5\%$ in adversarial accuracy of a baseline method, the DkNN-robustified CNN,  without appreciable loss of accuracy on clean data.    
\end{abstract}

\section{Introduction}
\label{sec: intro}
Primarily motivated by cognitive neuroscience, deep neural networks (DNNs) have reached impressive performance on various tasks \citep*{mehdipour2016comprehensive,zhao2019object,young2018recent}. The neuro-mimetic point of view has driven both the development of machine learning architectures and machine learning procedures. 
For example, early inspiration for the convolutional neural network (CNN) came from models of the visual cortex of the brain \cite{lindsay2020convolutional} and the popular machine learning frameworks of transfer learning lifelong learning are firmly rooted in cognitive psychology \cite{thrun2012learning, parisi2019continual}. However, the extraordinary performance of current DNNs on clean data still have a frustrating lack of robustness to small covariate shifts, e.g., adversarial attacks or drift of test data from their nominal training distributions. In the natural world such robustness is instantiated by another highly evolved biological system: the adaptive immune system. Motivated by analogies between robust machine learning and robustification mechanisms implemented by the immune system, we propose an immuno-mimetic framework for deep neural networks (Immuno-Net) that complements the neuro-cognitive perspective.  In this paper, we define the Immuno-Net framework and develop an artificial robust adaptive immune learning system (RAILS), which is inspired by the mechanisms of B-cell affinity maturation in the mammalian immune system. 

\begin{figure*}[h]
\centerline{\includegraphics[width=.93\textwidth]{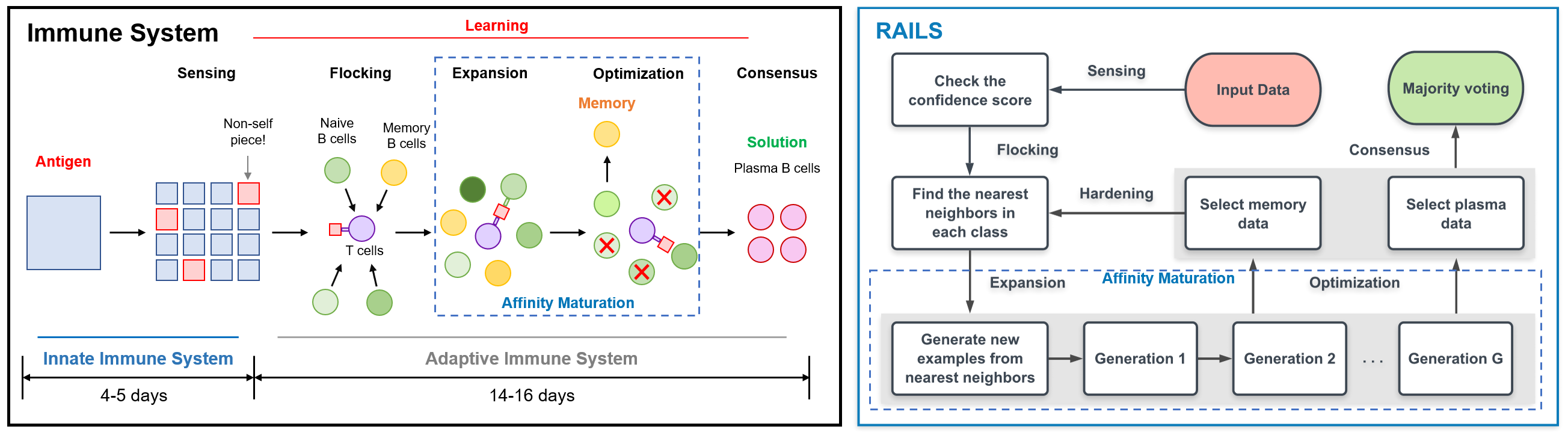}}
\vspace{-.12in}
\caption{Simplified biological immune system (left) and ImmunoNet RAILS computational workflow (right). Both systems are composed of a five-step process. We consider affinity maturation \cite{de2015dynamics} as the combination of expansion and optimization.}
\label{fig: bio_comp}
\end{figure*}

Similar to the human neural-cognitive system
\citep*{elsayedadversarial}, DNNs are vulnerable to small perturbations, e.g., from adversarial attacks \citep*{goodfellow2014explaining}.  Conversely, the mammalian immune system has the inherent capability to defend against a diverse set of attacks (bacterial, viral, fungal, and other infections) due to its built-in discrimator between self and non-self components  \citep*{farmer1986immune}. Through mechanisms of selection, hypermutation, and proliferation the adaptive immune system produces B-cells with increasing affinity to the external threat and is able to employ these solutions to further increase robustness by learning from multiple such attacks \citep*{mesin2016germinal, mesin2020restricted}. Furthermore, the immune system is able to further increase robustness by adaptively learning from multiple attacks \citep*{mesin2020restricted}.

The proposed Immuno-Net RAILS approach emulates the biological immune system {\em in silico}.


%
Our framework differs significantly from existing DNN robustification methods such as: outlier detection \citep*{metzen2017detecting,feinman2017detecting,grosse2017statistical}, and training-phase DNNs robustification \citep*{madry17,zhang2019theoretically,cohen2019certified} 
in that the immune system emulation provides a  complete end-to-end solution to the adversarial robustness problem. 

Specifically, we make the following contributions:
\begin{itemize}
\item Immuno-Net RAILS is the first robust adversarial defense framework for DNNs, based on the natural adaptive immune system. In particular:  (1) Immuno-Net RAILS emulates the principal mechanisms of the immune response to defend against novel attacks; (2) The learning rates exhibited by RAILS align with those of the immune system; and (3) Immuno-Net RAILS can further improve robustness by emulating adaptive learning (life-long learning) in the immune system.



\item We demonstrate that Immuno-Net RAILS improves adversarial robustness of the Deep k-Nearest Neighbor (DkNN) CNN  \citep*{papernot2018deep} by $5.62\%$, $12.5\%$, $10.32\%$ for the MNIST, SVHN, and CIFAR-10 datasets, respectively.
\end{itemize}

\section{Immune System to Computational System}

\subsection{Simplified Immune System}
The mammalian adaptive immune system has designed robustness into its biological architecture \citep*{rajapakse2011emerging}. The adaptive immune system is highly complex, so we simplify its learning process into five sub-processes: sensing, flocking, expansion, optimization, and consensus \citep*{cucker2007emergent,rajapakse2017emergence}. As shown in Fig.~\ref{fig: bio_comp}, \textit{sensing} of an attack leads to initial B-cells \textit{flocking} to lymph nodes, and forming temporary structures called germinal centers \citep*{de2015dynamics,farmer1986immune}. Germinal centers are populated in the \textit{expansion} and \textit{optimization} phases (together called affinity maturation), where a diverse set of B-cells bearing antigen-specific immunoglobulins divide symmetrically and asymmetrically with mutation and B-cells with  high affinity to the antigen are selected to the next generation  \citep{mesin2016germinal}. Memory B-cells are selected and stored within the evolving process, and can be used to defend against similar future attacks. In the \textit{consensus} phase, plasma B-cells are generated by selecting those B-cells that reach a threshold affinity against the foreign antigen. Plasma B-cells (along with their generated antibodies) represent the optimal solution of the adaptive immune response to a specific attack.

\subsection{From Immune Response to Computation}
Motivated by the simplified biological immune system, we propose a new immuno-mimetic learning strategy - the Immuno-Net Robust Adversarial Immune-inspired Learning System (Immuno-Net RAILS). 
The comparison in Fig.~\ref{fig: bio_comp} shows that both the biology in the immune system and the computational biology in Immuno-Net RAILS are composed of the same five-step process. Similar to the B-cell population growth during the clonal expansion, RAILS enlarges the population of candidates then selects high-affinity and moderate-affinity candidates to proliferate to predict the present inputs and defend against future attacks.

\begin{figure}[h]
\centerline{\includegraphics[width=.43\textwidth]{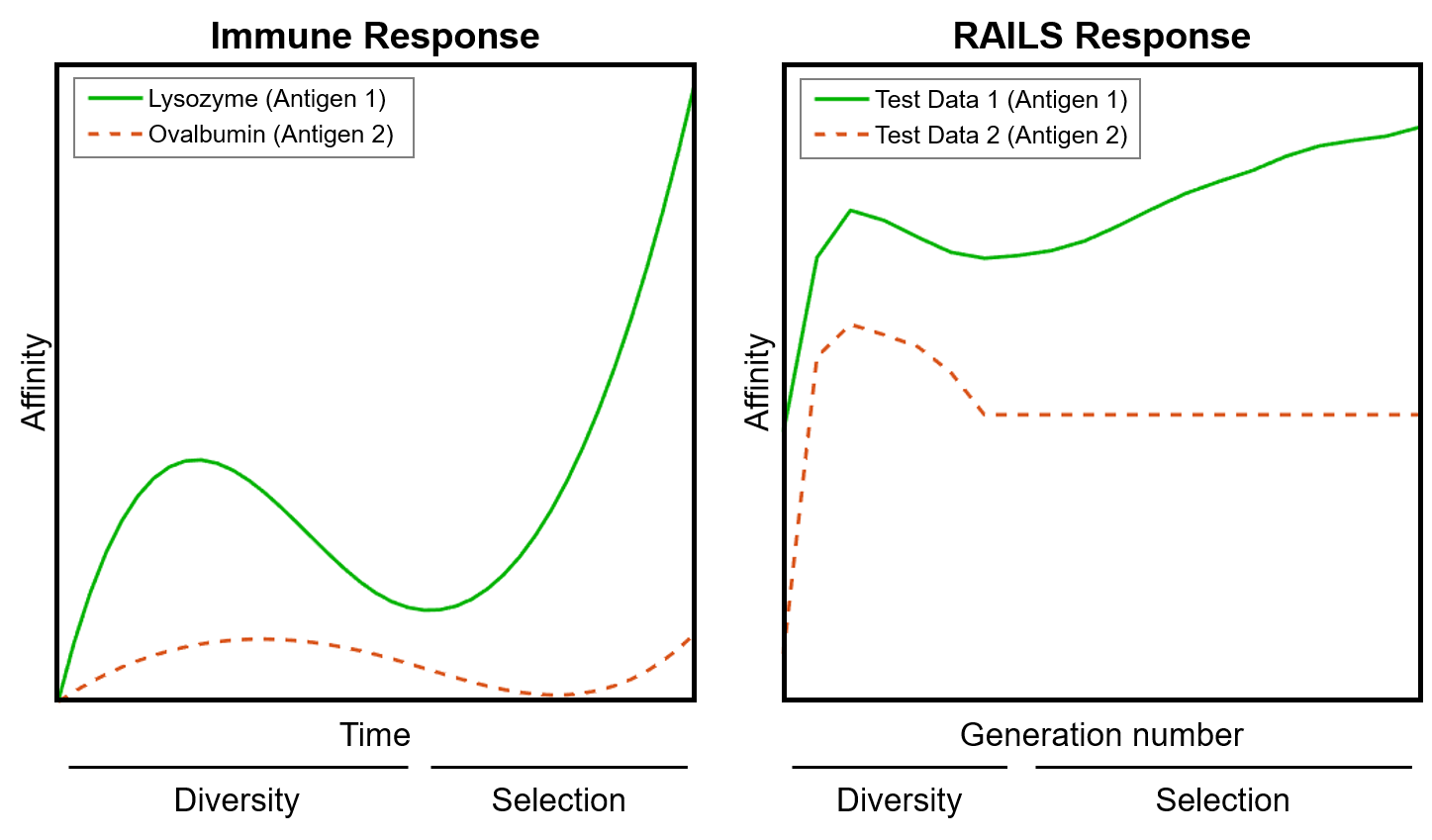}}
\vspace{-.05in}
\caption{Correspondence between learning rates of Immuno-Net RAILS computational biology model (right) and observed learning rates of {\em in vitro} B-cells in the natural immune system (left). The flatten region in the red curve (right) corresponds to Immuno-Net RAILS early stopping when there is no solution. The biological experiments were performed in the Rajapakse Lab.}
\label{fig: curves_bio_rails}
\end{figure}

Fig.~\ref{fig: curves_bio_rails} shows the learning curves of Immuno-Net RAILS and the immune system and demonstrates that Immuno-Net RAILS captures key properties of the immune system. The learning curves depict the change of affinity between the population of defenders (B-cells) and the threat (antigen) over time. Note that in the computational system, the antigen is the test data. The initial B-cells (initial candidates) are all from the flocking phase of response to antigen 1 (test data 1), which results in a mismatch to antigen 2 (test data 2), and thus cause the red affinity curves to remain relatively low. 
One can also see that both green curves meet a small drop at the early stage and monotonically increase thereafter. This {\rm two-phase learning process} happens due to a {\rm diversity vs. selection} trade-off;  diversity comes from the population size increasing and the randomness in generating new data points, while the selection arises from selection of high-affinity B-cells (newly generated data points).

\begin{algorithm}[H]
\caption{Immuno-Net RAILS}
\label{RAILS}
\begin{algorithmic}[1]
\REQUIRE Test data point $\mathbf x$; Training dataset $\din_{tr} = \{\din_{1}, \din_{2},\cdots,\din_{C}\}$; Number of Classes $C$; Model $M$ with feature mapping $f_l(\cdot)$, $l \in \mathcal{L}$; Affinity function $A$.\\
\noindent{\textbf{Sensing}}
\STATE{Check the threat score to detect the threat of $x$.}\\
\noindent{\textbf{Flocking}}
\FOR{$c = 1, 2, \dots, C$}
\STATE{In each layer $l \in \mathcal{L}$, find the k-nearest neighbors $\mathcal{N}_l^c$ of $\mathbf x$ in $\din_{c}$ by ranking the affinity score $A(f_{l}; \mathbf x_j, \mathbf x), \mathbf x_{j} \in \din_{c}$}
\ENDFOR\\
\noindent{\textbf{Affinity maturation: Expansion and Optimization}}
\STATE{\textbf{For} each layer $l \in \mathcal{L}$, \textbf{do}}
\STATE{Generate $\mathcal{P}^{(0)}$ through mutating each of $\mathbf x^{\prime} \in \textit{Union}(\{\mathcal{N}_l^1,\mathcal{N}_l^2,\cdots,\mathcal{N}_l^C\})$ $T/kC$ times.}
\FOR{$g = 1, 2, \dots, G$ }
\FOR{$t = 1, 2, \dots, T$}
\STATE{Select data-label pair $(\mathbf x_p, y_p)$ from $\mathcal{P}^{(g-1)}$ based on $\mathbf P_{g-1}$ = \textit{Softmax}($A(f_{l}; \mathcal{P}^{(g-1)}, \mathbf x)/\tau)$.}
\STATE{Pick all the data $S_{y_p}$ belonging to class $y_p$ in $\mathcal{P}^{(g-1)}$ and calculate $\mathbf P_{y_p} = \frac{\mathbf P_{g-1}(S_{y_p})}{\sum \mathbf P_{g-1}(S_{y_p})}$}
\STATE{Select another pair $(\mathbf x_{p}^{\prime}, y_p)$ and apply cross-over and mutation on the two pairs of data. Add the obtained offspring $\mathbf x_{\text{os}}$ to $\mathcal{P}^{(g)}$.}
\ENDFOR
\ENDFOR

\STATE{Calculate the affinity score $A(f_{l}; \mathcal{P}^{(G)}, \mathbf x), \forall l \in \mathcal{L}$}
\STATE{Select the top $5\%$ as plasma data $S_{p}^l$ and the top $25\%$ as memory data $S_{m}^l$ based on the affinity scores, $\forall l \in \mathcal{L}$}
\STATE{\textbf{end For}}\\
\noindent{\textbf{Consensus}}
\STATE{Obtain the prediction $y$ of $\mathbf x$ using the majority vote of the $S_{p} = \{S_{p}^1, S_{p}^2, \cdots, S_{p}^{|\mathcal{L}|}\}$}
\STATE {\bfseries Output:} $y$, the memory data $S_{m} = \{S_{m}^1, S_{m}^2,\cdots,S_{m}^{|\mathcal{L}|}\}$
\end{algorithmic}
\end{algorithm}

\section{Immuno-Net RAILS Computations}\label{sec: RAILS}

In this section, we describe the implementation  of Immuno-Net RAILS as {\em in silico} analogs to flocking through consensus phases in immune response (Fig. \ref{fig: bio_comp}). 
Given the mapping from input to layer $l$, $f_l: \mathbb{R}^{d} \rightarrow \mathbb{R}^{d^{\prime}}$ and $\mathbf x_1, \mathbf x_2 \in \mathbb{R}^{d}$, we first define the affinity $A(f_l;\mathbf x_1, \mathbf x_2)$ between $\mathbf x_1$ and $\mathbf x_2$ on layer $l$. The affinity can be measured in different ways, e.g, inverse $\ell_p$ distance, cosine similarity, or inner product. Here we use the negative Euclidean distance measurement, i.e., $A(f_l;\mathbf x_1, \mathbf x_2) = -\|f_l(\mathbf x_1) - f_l(\mathbf x_2)\|_2$. Algorithm~\ref{RAILS} depicts Immuno-Net RAILS' five-step workflow. We explain each step in detail in the subsequent section.

\subsection{Five-Step Workflow}

\paragraph{Sensing.} This step emulates the immune system's self/non-self detection and performs the initial discrimination between perturbed inputs and clean inputs. The sensing step can effectively prevent Immuno-Net RAILS from becoming overwhelmed by false positives, i.e., treating clean inputs as adversarial inputs. This is implemented using an outlier detection procedure, for which there are several different methods available \citep*{feinman2017detecting,xu2017feature}. 
The sensing stage provides a confidence score of the architecture and does not affect Immuno-Net RAILS's predictions. 

\paragraph{Flocking.} Similar to the B-cells flocking to lymph nodes, Immuno-Net RAILS leverages flocking to provide an initial population for the affinity maturation process.  At the implementation level, flocking is used to find the k-nearest neighbors that have the highest initial affinity score to the input data in each selected layer and from each class. We construct kNN sets $\mathcal{N}_l^c $ independently for each class $c$, thereby ensuring that the initial population for every class is balanced.
\begin{align}\label{eq: knn}
    \begin{array}{ll}
\mathcal{N}_l^c = \{(\hat{\mathbf x}, y_c)|R_c(\hat{\mathbf x}) \le k, (\hat{\mathbf x}, y_c) \in \din_c\} \\ \text{Given}\\ A(f_l; \mathbf x_i^c, \mathbf x) \le A(f_l; \mathbf x_j^c, \mathbf x) \Leftarrow R_c(i) > R_c(j)\\ \forall c \in [C], l \in \mathcal{L},  \forall i, j \in [n_c],
    \end{array}
\end{align}
where $\mathcal{L}$ is the set of the selected layers. $\mathbf x$ denotes the input. $\din_c$ represents the training data from class $c$ with the size $|\din_c|=n_c$. $R_c: [n_c] \rightarrow [n_c]$ is a ranking function that sorts the indices based on the affinity score. The candidates for flocking might purely come from the original training data or include the memory database which will be clear in the optimization stage introduction. 

\paragraph{Expansion and Optimization.} The expansion step starts from the initial population, and generates new examples (which are offspring) from the existing examples (which are parents) in the population. The ``initial B-cells'' are nearest neighbors found by \eqref{eq: knn} in the flocking step, where there are $kC$ data points in total. We then enlarge the population size to $T$ by copying each nearest neighbor $\frac{T}{kC}$ times with random mutations. We call the fist $T$ size population as the $0$-th generation. We now introduce the rule of generating new generations. Let $\mathcal{P}^{(g-1)}=[\mathbf x_p(1), \mathbf x_p(2), \cdots, \mathbf x_p(T)] \in \mathbb{R}^{d \times T}$ denote the population in the $(g-1)$-st generation. Then the candidates for generating the $g$-th generation are selected by
\begin{align}\label{eq: random}
    \begin{array}{ll}
\mathcal{\hat{P}}^{(g-1)} = \mathcal{P}^{(g-1)} Z_{g-1},
    \end{array}
\end{align}
where $Z_{g-1} \in \mathbb{R}^{T \times T}$ is a binary selection matrix whose columns are independent and identically distributed draws from Mult$(1, \mathbf P)$, the multinomial distribution with selection probability vector $\mathbf P\in [0,1]^T$. The selection probability for each candidate is obtained through a softmax function.
\begin{align}\label{eq: prob}
    \begin{array}{ll}
\mathbf P(\mathbf x_i) &= Softmax(A(f_{l}; \mathbf x_i, \mathbf x)/\tau) \\&= \frac{\exp{(A(f_{l}; \mathbf x_i, \mathbf x)/\tau)}}{\sum_{\mathbf x_j \in S}\exp{(A(f_{l}; \mathbf x_j, \mathbf x)/\tau)}},
    \end{array}
\end{align}
where $S$ is the set of data points and $x_i \in S$. $\tau>0$ is the sampling temperature that controls sharpness of the softmax operation. 

We further leverage \textit{cross-over} and \textit{mutation} operations to increase candidate diversity. In Immuno-Net RAILS, we select two parents from the selected candidates \eqref{eq: random} that share the same label, and apply the cross-over operator to combine the two parents together. The cross-over combination is implemented by selecting at random each of its elements (pixels) from the corresponding element of either parent1 or parent2 based on their affinity score. After the combination, mutation is leveraged to generate the offspring $\mathbf x_{\text{os}}$ that belongs to the $(g-1)$-st generation. This operation randomly and independently  mutates the data with probability $\rho$, adding uniformly distributed noise in the range $[-\delta_{\text{max}},-\delta_{\text{min}}] \cup [\delta_{\text{min}},\delta_{\text{max}}]$. Given examples lie in $[0,1]^d$, the resulting perturbation vector is subsequently clipped to satisfy the domain constraint. 



After multiple generations, RAILS selects generated examples with high-affinity scores to be plasma data, and examples with moderate-affinity scores are saved as memory data. Similar to plasma B-cells, plasma data represents an optimal solution to the current attack. Memory data can be seen as an analog to memory B-cells, which help warm-start defense responses to similar attacks in future. We set the threshold to be $0.05$ and $0.25$ for selecting plasma data and memory data, respectively, i.e, top $5\%$ of final generation will become plasma data, and top $25\%$ will become memory data. $S_{p} = \{S_{p}^1, S_{p}^2, \cdots, S_{p}^{|\mathcal{L}|}\}$ denotes the plasma data we selected from each layer. The memory data is denoted by $S_{m} = \{S_{m}^1, S_{m}^2,\cdots,S_{m}^{|\mathcal{L}|}\}$. In this paper, we will only focus on static learning, i.e., only leveraging the plasma data to make predictions.


\paragraph{Consensus.} All the new examples generated in each generation are associated with labels that are inherited from their parents. The consensus step uses majority voting of the plasma data for prediction of the class label of input $\mathbf x$.

\section{Experimental Results}
We compare Immuno-Net RAILS to standard Convolutional Neural Network Classification (CNN) and Deep k-Nearest Neighbors Classification (DkNN) \cite{papernot2018deep} on the MNIST \cite{lecun1998gradient}, SVHN \cite{netzer2011reading}, and CIFAR-10 \cite{Krizhevsky2009learning}. 
In addition to the benign test examples, we also generate the same amount of adversarial examples using PGD attack \cite{madry17}. The attack strength is $\epsilon = 60/8/8$ for MNIST, SVHN, and CIFAR-10 by default. The performance will be measured by standard accuracy (SA) evaluated using benign test examples and robust accuracy (RA) evaluated using the adversarial test examples.

Immuno-Net RAILS leverages static learning to make the predictions. The results are shown in Table~\ref{tab: acc_overall}. One can see that RAILS delivers a $5.62\%/10.32\%/12.5\%$ improvements in RA over DkNN without appreciable loss of SA on the three datasets. 

\begin{table}[h]
\vspace{-0.13in}
\begin{center}
\caption{
SA/RA performance of Immuno-Net RAIL (RAILS) versus CNN and DkNN. Immuno-Net RAILS obtains higher RA for all three datasets.}
\label{tab: acc_overall}
\vspace{.02in}
\resizebox{0.42\textwidth}{!}{
\begin{tabular}{lc|c|c}
\hline
\hline 
& & SA & RA \\
\hline 
MNIST &\bf{RAILS (ours)} & 97.95\% & \bf{76.67\%}  \\

($\epsilon=60$) &CNN & \bf{99.16\%} & 1.01\% \\

& DkNN & 97.99\% & 71.05\%   \\
\hline 
SVHN  &\bf{RAILS (ours)}  & 90.62\% & \bf{48.26\%} \\

($\epsilon=8$) &CNN & \bf{94.55\%} & 1.66\% \\

&DkNN & 93.18\% & 35.7\% \\
\hline 
CIFAR-10 &\bf{RAILS (ours)} & 82\% & \bf{52.01\%}\\

($\epsilon=8$) &CNN & \bf{87.26\%} & 32.57\% \\

&DkNN & 86.63\% & 41.69\% \\

\hline
\hline
\end{tabular}}
\end{center}
\end{table}

\vspace*{-\baselineskip}
\section{Conclusion}
Inspired by the immune system, we proposed a new defense framework for deep learning models. 
Immuno-Net RAILS is able to emulate biological processes with evolutionary programming that model natural selection and affinity maturation of B-cells in the natural immune system. 
Our experiments show that the Immuno-Net RAILS learning curve mimics the diversity-selection learning phases observed in {\em in vitro} B-cell affinity maturation experiments in the Rajapakse lab. 

In future work, we will explore the mechanisms of the immune system's adaptive learning (life-long learning) and covariate shift adjustment, which will be  consolidated into our computational framework. 

\section*{Acknowledgement}
We thank Alanawaz Rehemtulla and Sivakumar Jeyarajan for helpful discussions and guidance on the adaptive immune system. This work is supported by the Guaranteeing AI Robustness against Deception (GARD) program from DARPA/I2O.


\bibliography{refs_adv}





\end{document}